\documentclass[runningheads]{llncs}
\usepackage{graphicx}
\usepackage{xcolor}
\usepackage{booktabs}
\usepackage{caption}
\usepackage{subcaption}
\usepackage{amsmath}
\begin{document}
\title{Predicting COVID-19 Patient Shielding: A Comprehensive Study }
\titlerunning{{Predicting COVID-19 Patient Shielding}}
%
\author{Vithya~Yogarajan\orcidID{0000-0002-6054-9543} \and
Jacob~Montiel\orcidID{0000-0003-2245-0718} \and
Tony~Smith\orcidID{0000-0003-0403-7073} \and Bernhard~Pfahringer\orcidID{0000-0002-3732-5787}}

%
\authorrunning{Yogarajan et al.}

\institute{Department of Computer Science, University of Waikato, New Zealand 
\email{vy1@students.waikato.ac.nz}}
\maketitle              
\begin{abstract}

There are many ways machine learning and big data analytics are used in the fight against the COVID-19 pandemic, including predictions, risk management, diagnostics, and prevention. This study focuses on predicting COVID-19 patient shielding --- identifying and protecting patients who are clinically extremely vulnerable from coronavirus. This study focuses on techniques used for the multi-label classification of medical text. Using the information published by the United Kingdom NHS and the World Health Organisation, we present a novel approach to predicting COVID-19 patient shielding as a multi-label classification problem. We use publicly available, de-identified ICU medical text data for our experiments. The labels are derived from the published COVID-19 patient shielding data. We present an extensive comparison across 12 multi-label classifiers from the simple binary relevance to neural networks and the most recent transformers. To the best of our knowledge this is the first comprehensive study, where such a range of multi-label classifiers for medical text are considered. We highlight the benefits of various approaches, and argue that, for the task at hand, both predictive accuracy and processing time are essential.

\end{abstract}
\section{Introduction}

The Coronavirus disease 2019 (COVID-19) pandemic has presented a considerable challenge to the world health care system, and the management of COVID-19 is an ongoing struggle. The ability to identify and protect high-risk groups is debated by the scientific community~\cite{ioannidis2021precision}. COVID-19 patient shielding refers to identifying and protecting patients who are clinically extremely vulnerable from coronavirus. Patients in these categories include those people who have been identified by health professionals before the pandemic as being clinically extremely vulnerable, and those identified through the COVID-19 Population Risk Assessment model~\cite{Cliftm3731}. Such patients present with co-morbidity~\cite{Cliftm3731} and hence the need to use multi-label classification techniques. The clinical risk prediction model (QCOVID)~\cite{Cliftm3731} was developed by the United Kingdom NHS based on a cohort study of the population using data from over 6 million adults. It provides guidelines for patients who are considered high risk (clinically extremely vulnerable); for example, patients with a certain type of cancer, patients who have had organ transplants, and patients having a serious heart condition while also being pregnant. 

\begin{figure}[t]
    \centering
    \includegraphics[width=0.97\textwidth]{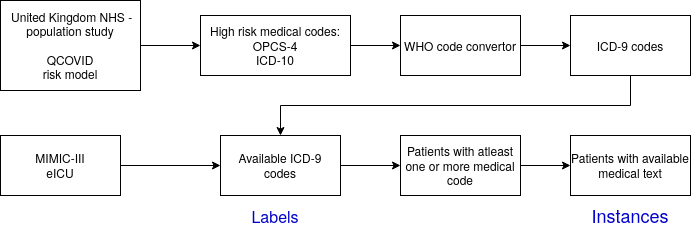}
    \caption{Flow chat forming labels and instances for multi-label classification of predicting COVID-19 patient shielding. }
    \label{fig:covid}
    \vspace{-1em}
\end{figure}

QCOVID provides information on medical codes which can be used in hospital databases to identify patients who are at high risk and fall under the criteria of COVID-19 shielding. Due to privacy and legal issues, obtaining current patient records from hospitals is not
possible~\cite{datashare}. However, electronic health records (EHRs) from publicly available data such as Medical Information Mart for Intensive Care (MIMIC-III)~\cite{johnson2016mimic,goldberger2000physiobank} and electronic Intensive Care Unit (eICU)~\cite{goldberger2000physiobank,PollardSD2018} present a realistic set of EHRs obtained and de-identified from hospitals in the United States. This research presents a novel approach of considering COVID-19 patient shielding as a multi-label problem (see Figure~\ref{fig:covid} for details). Multi-label classification assigns a set of labels to an instance. From a collection of labels, each record will be assigned relevant medical codes/labels. Using the medical codes presented in QCOVID and the code conversions presented by the WHO, we identify and predict patients with one or more high-risk medical codes associated with EHRs in MIMIC-III and eICU. 

This research presents an extensive comparative study across 12 multi-label classifiers starting from the simplest binary relevance with logistic regression to neural networks and several transformers. We highlight the advantages and disadvantages of these classifiers when applied to predicting COVID-19 patient shielding. We present overall predictive accuracy as well as label F1-scores where each label is a high-risk COVID-19 medical code. We show that, in addition to the overall predictive accuracy, processing time plays a big role in decision making.\footnote{The code to recreate the experiments and evaluations described in this paper is accessible at: https://github.com/vithyayogarajan/COVID-19-Patient-Shielding}  

The contributions of this work are:
\begin{enumerate}
    \item we present a novel approach of considering predicting COVID-19 patient shielding as a multi-label problem; 
    \item we analyse the effectiveness of 12 different multi-label classifiers including several neural networks and transformers for predicting medical codes associated with COVID-19 patient shielding from medical text;
    \item we show both overall prediction accuracy and processing time plays a role in selecting a multi-label approach;
\end{enumerate}

\section{Related Work}

The advancement of machine learning-based predictions motivates researchers to consider ways to help in the fight against COVID-19, and there are numerous ways in which this can be achieved~\cite{alimadadi2020artificial}. The year 2020 has seen many examples of machine learning approaches applied to various aspects of the COVID-19 pandemic. Some examples of machine learning approaches include:
\begin{itemize} 
\item a hybrid machine learning technique for pandemic predictions based on data from Hungary~\cite{pinter2020covid}
\item a deep convolutional neural network model named CoroNet for predicting COVID-19 from chest X-ray images\cite{khan2020coronet}
\item support vector machines for predicting radiological findings consistent with COVID-19 from radiology text reports~\cite{lopez2020covid}
\item classifying clinical reports using logistic regression and multinomial Naive Bayes~\cite{khanday2020machine}
\item spectrometric data analysis for a diagnosis based study~\cite{delafiori2021covid}
\end{itemize}

These examples are only the beginning of the endless possibilities of machine learning approaches for the COVID-19 pandemic. Several machine learning methods are used for outbreak predictive studies worldwide to make decisions and enforce control measures. The systems mentioned above are only a small subset of examples of such studies. This reflects a need for extensive comparison among machine learning models, especially for predictions from medical text. 

Yogarajan et al. (2020)~\cite{yogarajan2020seeing} show domain-specific skip-gram pre-trained fastText embeddings perform better that general text pre-trained embeddings. Hence, 100 and 300-dimensional domain-specific fastText pre-trained embeddings~\cite{yogarajan2020,yogarajan2020seeing} are used for this research. A collection of neural networks and transformer models are chosen for the predictive study in this research, based on literature where such models have been used for multi-label medical code prediction and have achieved state-of-the-art results~\cite{yogarajan2021trans,mullenbach2018explainable,moons2020comparison}.

\section{Data and Labels}\label{sec:data}

\begin{table}[t!]
    \centering
    \caption{Summary of Data for COVID-19 patient shielding, for EHRs with available relevant medical codes is included. Percentage frequency of occurrences of labels for MIMIC-III and eICU is also presented.  }
    \label{tab:data}
     \resizebox{\linewidth}{!}{
    \begin{tabular}{l|l}
    \hline
    \noalign{\smallskip}
    MIMIC-III     & eICU  \\
    \noalign{\smallskip}
    \hline
    \noalign{\smallskip}
    Critical care units at the & The Philips eICU program.  \\ 
    Beth Israel Deaconess Medical  & Admitted to ICUs in 2014 and 2015 \\ 
    Center between 2001 - 2012. &across the United States. \\ & \\
    Free-form medical text. Eg: & Semi-structured medical text. Eg: \\ 
    \protect\it{``59M w HepC cirrhosis c/b grade } &\protect\it {``infectious diseases $|$ medications $|$}   \\ 
     \protect\it {I/II esophageal varices and portal } & \protect\it {therapeutic antibacterials $|$ cardio $|$}\\  
      \protect\it{ gastropathy, p/w coffee-ground''} &\protect\it {inotropic agent $|$ norepinephrine'' }\\ & \\
    35,458 hospital admissions with      & 34,387 patients with EHRs.  \\
    discharge summaries. & \\  Text length: 60 - 9500 tokens. & Text length: 50 -
1400 tokens.\\ & \\
    42 medical codes (labels) & 25 medical codes (labels) \\
   \includegraphics[width=0.49\textwidth,height=3.2cm]{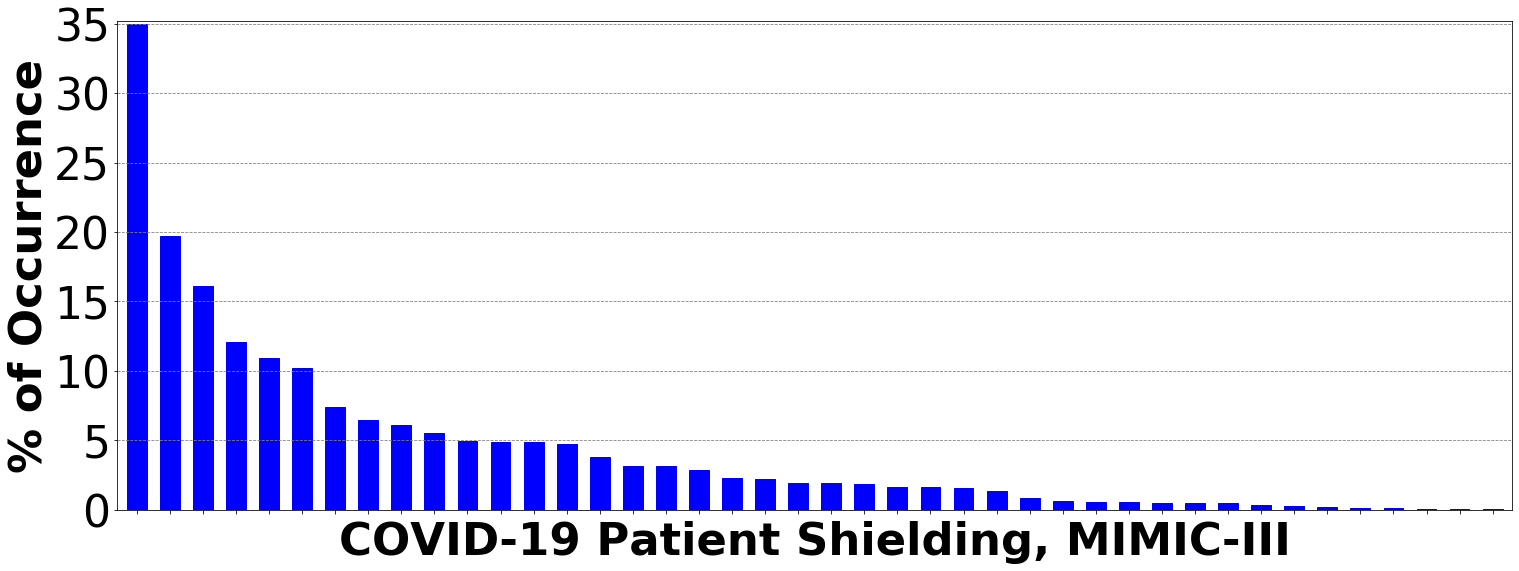}
   &  \includegraphics[width=0.49\textwidth,height=3.2cm]{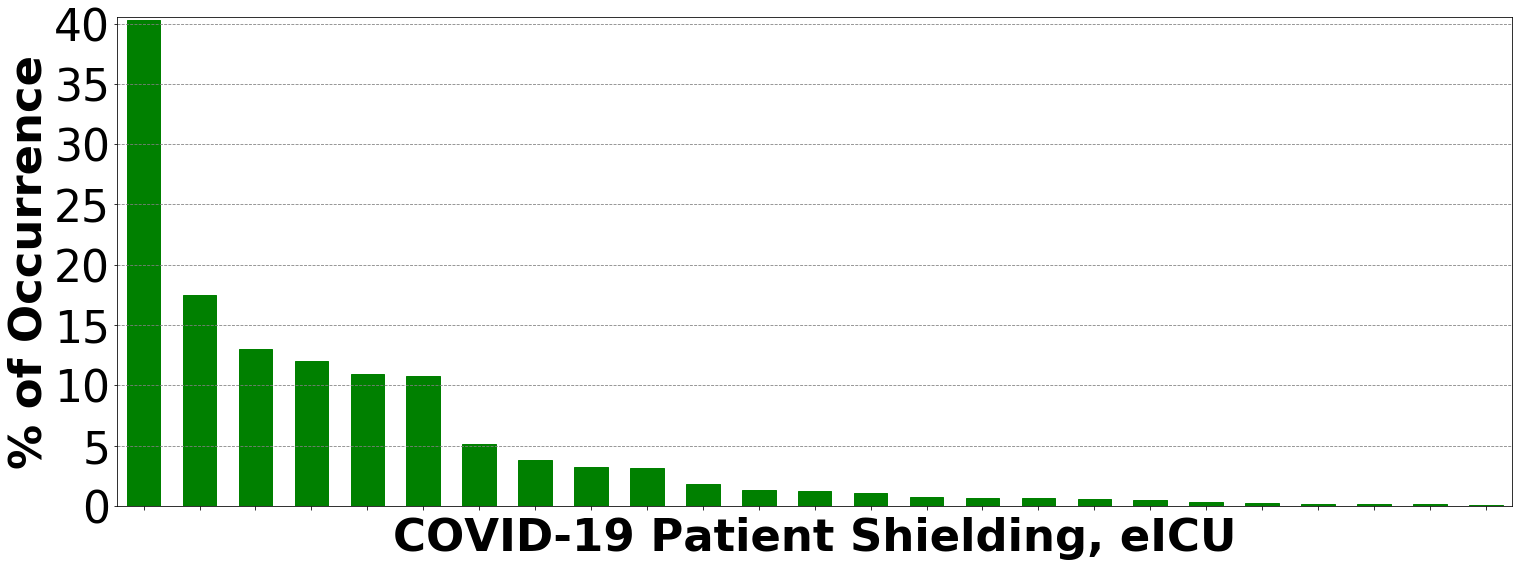}\\
   ICD-9 code examples: & ICD-9 code examples: \\
  Code 285 (35\%) Other and unspecified anemias & Code 491 (40.3\%) Chronic bronchitis \\
  Code 996 (16\%) Complications peculiar to  & Code 288 (17.5\%) Diseases of white blood cells \\
  certain specified procedures & \\
    \noalign{\smallskip}
    \hline
    \end{tabular}}
\end{table}

ICD (standards for International statistical Classification of Diseases and related health problems) codes are used to classify diseases, symptoms, signs, and causes of diseases. For this research, we use ICD-9 because the available data pnly includes labels for ICD-9 codes. A summary of MIMIC-III and eICU, with details of number of instances and labels, label frequencies and examples of EHRs and ICD-9 codes are presented in Table~\ref{tab:data}.

\section{Multi-Label Classifiers}

\begin{figure}[t!]
      \subfloat[]{ 
        \label{fig:A}
        \includegraphics[height=5.7cm]{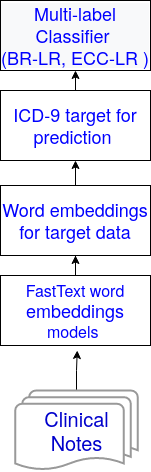}
    } 
 \hfill  
      \subfloat[]{ 
        \label{fig:B}
        \includegraphics[height=5.7cm]{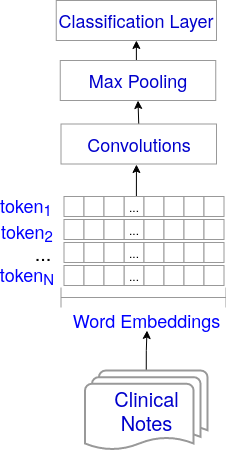}
    } 
\hfill  
    \subfloat[]{ 
        \label{fig:C}
       \includegraphics[height=5.7cm]{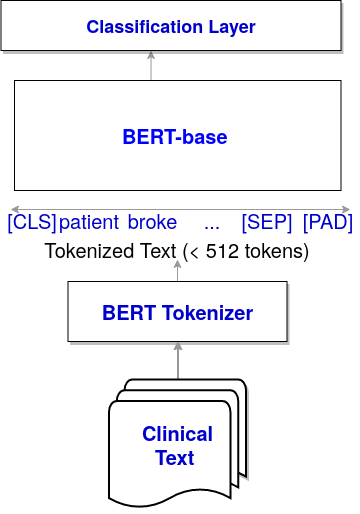}
    }
    \caption{(\protect\subref{fig:A}) Flowchart for traditional multi-label classifiers; (\protect\subref{fig:B}) CNNText~\cite{kim2014convolutional} architecture; and (\protect\subref{fig:C}) Example of BERT-base model for multi-label classification. } \label{fig:trans}%
    \vspace{-1em}
\end{figure}

\begin{table}[t!]
    \centering
     \caption{Summary of Transformer Models. Except for general text pre-trained models Longformer and TransformerXL, all other models included in this research can only handle a sequence length of at most 512 tokens. Continuous training approach would initialize with the standard BERT model, pre-trained using Wikipedia and BookCorpus. It then continues the pre-training process with masked language modelling and next sentence prediction using domain-specific data. Similarly for BioMed-RoBERTa.  }
    \label{tab:summary}
     \resizebox{\linewidth}{!}{
    \begin{tabular}{lll}
    \noalign{\smallskip}
    \hline
    \noalign{\smallskip}
         Transformers &  Training Data\quad \quad & Model Details  \\ \noalign{\smallskip}\hline
         
         BERT~ \cite{DBLP:journals/corr/abs-1810-04805} & {Books + Wiki}  & 12-layers with a hidden size of 768, 12 self-attention\\
         & &  heads,  110M parameter neural network architecture.   \\ & & \\
         ClinicalBERT~\cite{alsentzer2019publicly}  &  MIMIC-III & continuous training of BERT-base \\ & & \\
         PubMedBERT~\cite{gu2020domain}  & PubMed & BERT-base model architecture \\ 
         & & Trained from scratch  \\ & & \\
         RoBERTa-base~\cite{DBLP:journals/corr/abs-1907-11692} & Web crawl & robustly optimised BERT approach \\ 
         & &  improved training methodology   \\ & & \\
         BioMed-RoBERTa~\cite{domains}  &Scientific papers \quad & continuous training of RoBERTa-base \\ & & \\  \hline & & \\
         TransformerXL~\cite{dai2019transformer}   &\multicolumn{2}{l}{Can handle documents with $> 512$ tokens. Models longer-term dependency }   \\
               & \multicolumn{2}{l}{by combining recurrence and relative positional encoding.} \\ & \multicolumn{2}{l}{Training Data: general text including Wiki}\\ & & \\
         Longformer~\cite{beltagy1904longformer}   &\multicolumn{2}{l}{Can handle documents with $> 512$ tokens. }  \\
         & \multicolumn{2}{l}{Provides computational and memory efficiency.}\\
          & \multicolumn{2}{l}{Training Data: Books + Wiki + Realnews + Stories}\\\noalign{\smallskip} \hline
 
    \end{tabular}}
   \vspace{-1em}
\end{table}

The first and simplest multi-label classification algorithm is  binary relevance (BR)~\cite{godbole2004discriminative}. A separate binary classification model is created for each label, such that any text with that label is a positive instance, 
and all other records form the negative instances.
BR models make their predictions independently. However, for multi-label problems where there is a strong correlation between labels, a model could benefit from the result of another label when making its predictions. BR models can be `chained' together into a sequence such that the predictions made by earlier classifiers are made available as additional features for the later classifiers. Such a configuration is called a classifier chain (CC)~\cite{JesseRead2011}.  Ensembles of classifier chains (ECC) built with diverse random chaining orders can help mitigate the effect of chaining order. In this research, logistic regression (LR) is used as the base classifier for both BR and ECC. Figure~\ref{fig:A} provides a flow chart for BR and ECC, where document embeddings for each clinical document are used as features. Document level embeddings are computed by obtaining the vector sum of the embeddings for each word in the document and then normalising the sum to have length one, to ensure that documents of different lengths have representations of similar magnitudes.

CNNText~\cite{kim2014convolutional} combines one-dimensional convolutions with a max-over-time pooling layer and a fully connected layer. Each word or token in the clinical document is mapped to its $k$-dimensional embeddings using the lookup table for a given pre-trained embeddings model. A new feature is produced using the convolution operation, and max-over-time pooling is applied over the feature map to capture the most important feature value. Multiple filters are used with varying window sizes. The final prediction is made by computing a weighted combination of the pooled values and applying a sigmoid function. A simple architecture of CNNText is presented in Figure \ref{fig:B}. 

Gated Recurrent Units (GRU)~\cite{cho2014gru} are a type of recurrent neural networks which are designed to better handle vanishing gradients. Bidirectional GRU (Bi-GRU) considers both sequences from left to right and the reverse order. As with CNNText, clinical documents are mapped to their embeddings, and a sigmoid activation function is used. 

Mullenbach et al. (2018)~\cite{mullenbach2018explainable} present Convolutional Attention for Multi-Label classification (CAML), which achieves SOTA results for predicting ICD-9 codes from MIMIC III data~\cite{moons2020comparison}. CAML combines convolution networks with an attention mechanism. Simultaneously, a second module is used to learn embeddings of the descriptions of ICD-9 codes to improve the predictions of less frequent labels. A regularising objective with a trade-off hyperparameter was added to the loss function of CAML. This variant is called Description Regularized-CAML (DR-CAML)~\cite{mullenbach2018explainable}.

This research includes several variations of the most recent developments of NLP transformers~\cite{vaswani2017attention}. In this paper, we perform transfer learning of pre-trained transformer models by fine tuning for predicting COVID-19 patient shielding. All parameters are fine-tuned end-to-end.  A summary of Transformer models used in this research is presented in Table~\ref{tab:summary}. BERT-base variations and RoBERTa-base variations can only handle a maximum sequence length of 512, whereas Longformer and TransformerXL can handle longer sequences. Continuous training of transformers refers to existing general text pre-trained models that are further trained  with a masked language model and next sentence prediction using domain-specific data. The vocabulary is the same as the original BERT model, which is considered a disadvantage for domain-specific tasks~\cite{gu2020domain}. Alternatively, training from scratch indicates that the models are pre-trained with domain-specific data only and the vocabulary of these models is domain-specific. Figure~\ref{fig:C} provides an example of a BERT-base model.

\section{Experimental Setup}

Neural network models presented in this research are implemented using PyTorch. All evaluations were done using sklearn metrics and the transformer implementations are based on the open-source PyTorch-transformer repository. For multi-label classification MEKA~\cite{read2016}, an open-source Java system specifically designed to support multi-label classification experiments, is used for BR and ECC. Domain-specific fastText embeddings~\cite{yogarajan2020,yogarajan2020seeing} of 100-dimensions are used for neural networks and 300-dimensions for BR and ECC.

Transformer models were fine-tuned on all layers without freezing. Adam~\cite{kingma2014adam}, an adaptive learning rate optimisation algorithm, is used as the optimiser for all neural networks. A non-linear sigmoid function $f(z) = \frac{1}{1+e^{-z}}$, is used as the activation function. All experiments using ECC and BR are validated through 10-fold cross-validation. Due to resource restrictions, all neural network results, including transformers, use random seed train-test hold-out set validation where the results are averaged over three runs. Experiments for neural networks and transformers were run on a 12 core Intel(R) Xeon(R) W-2133 CPU @ 3.60GHz, and a GPU device GV100GL [Quadro GV100]. Experiments for BR and ECC were run on a 4 core Intel i7-6700K CPU @ 4.00GHz with 64GB of RAM.

For MIMIC-III data, discharge summaries are used as the free-form medical text and each hospital admission is treated as an instance. For ECC and BR, all available discharge summaries for each hospital admission are treated as one document, and 300 dimensional features are obtained from fastText pre-trained domain-specific embeddings. For embedding-based neural networks CNNtext, BiGRU, CAML and DRCAML, the pre-processed tokens from discharge summaries are truncated to 3,000 words. For BERT and RoBERTa based transformers, the input sequence is truncated to 512 tokens, and for Longformer and TransformerXL the input sequence is truncated to 3,000 and 3,072 tokens respectively. Discharge summaries were pre-processed by removing tokens that contain non alphabetic characters and special characters, and tokens that appear in fewer than three training documents. The extensively pre-processed eICU data was used `as is' for all experiments.   

\section{Results}

\begin{table}[tph!]
    \centering
\caption{Comparison of micro-F1 and macro-F1 of 
ICD-9 codes
for various multi-label classifiers for both MIMIC-III and eICU data. Time required per run is also presented. Bold is used to indicate the best results among the groups, and underline is used for overall best results. Results are averaged over three runs for neural networks and transformers. 10-fold cross validation is used for BR-LR and ECC-LR.}\label{tab:res_overall}
     \resizebox{\linewidth}{!}{
    \begin{tabular}{lrrrrrrr}
   \hline\noalign{\smallskip}
    & \multicolumn{3}{c}{MIMIC-III}  &\quad \quad& \multicolumn{3}{c}{eICU}  \\
Classifiers&Micro -F1&Macro-F1& Total Time &&Micro -F1&Macro-F1& Total Time\\
&&&(per run)&&&&(per run)\\\noalign{\smallskip}\hline\noalign{\smallskip}  
BR-LR & 0.39&0.26 & \textbf{12 min} & &  \textbf{0.54}&\textbf{0.28} & \textbf{7 min}\\
ECC-LR & \textbf{0.45}&\textbf{0.27} & 38 min & &0.51 &\textbf{0.28} &  34 min\\
& & & & & & & \\
CNNText   & 0.58 &\textbf{0.42} & 46 min & &0.59 & \textbf{0.36} & 45 min \\
BiGRU   & 0.59 &0.31 & 216 min & &0.59 &0.35 &210 min\\
CAML   &  \textbf{0.61} &0.40 & 49 min & & 0.60 & 0.32&48 min\\
DRCAML   & 0.60 &0.39 & 64 min & &\textbf{0.61} &0.32 & 60 min \\
& & & & & & & \\
BERT-base & 0.50 & 0.44 &10 hr & &0.60 & 0.36 & 11 hr \\
ClinicalBERT & 0.51 &0.45 &16 hr & &0.60 &0.36  & 11 hr \\
BioMed-RoBERTa &0.53 &0.45 & 12 hr & &0.61 &0.37 &11 hr \\
PubMedBERT &0.54 &0.48 & 16 hr& & \underline{\textbf{0.64}}&0.39 & 14 hr \\
Longformer &0.58 &0.50 & 82 hr & &0.61 &\underline{\textbf{0.40}} & 49 hr \\
TransformerXL & \underline{\textbf{0.65}}& \underline{\textbf{0.51}}& 206 hr & &0.63 &\underline{\textbf{0.40}} & 53 hr\\
\noalign{\smallskip}\hline
\end{tabular}}

\end{table}

\begin{figure}[bhp!]
    \begin{subfigure}[b]{\textwidth}
        \centering
        \includegraphics[width=\textwidth,height=2.8cm]{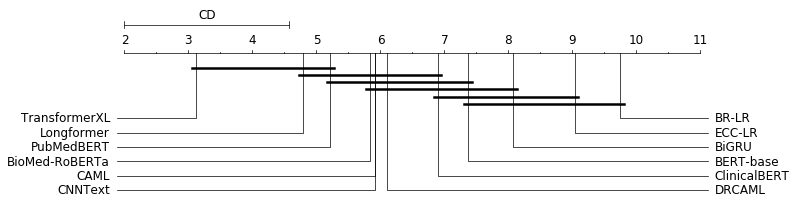}
        \caption{Data: MIMIC-III}
    \end{subfigure}
    \begin{subfigure}[b]{\textwidth}
        \centering
        \includegraphics[width=\textwidth,height=2.8cm]{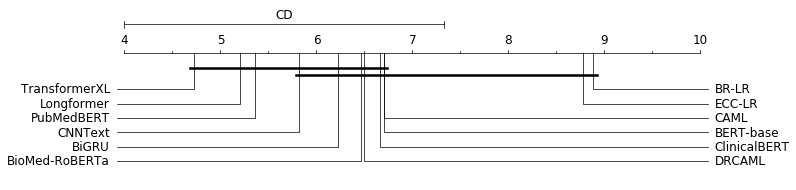}
        \caption{Data: eICU}
    \end{subfigure}
    \caption{Critical difference plots of label F1 scores for COVID-19 patient shielding with MIMIC-III (top) and eICU (bottom).  Nemenyi post-hoc test (95\% confidence level), identifying statistical differences between multi-label classifiers presented in Table \ref{tab:res_overall}. Critical difference is calculated for individual label F1 scores. }
    \label{fig:18_cd_eicu_bert}
    \vspace{-4pt}
\end{figure}
      
Table~\ref{tab:res_overall} presents a summary of overall micro and macro F1 scores for 12 multi-label classifiers and the total time per run of each experiment for both MIMIC-III and eICU data. Figure~\ref{fig:18_cd_eicu_bert} presents critical difference plots based on the label F1 scores for MIMIC-III and eICU to help identify the statistical difference between the multi-label classifiers. Compared to the basic BR-LR, which requires the least time for experiments, there is a significant improvement in overall micro and macro F1 measures for MIMIC-III data. This observation is also reflected on the critical difference plots where BR-LR is the classifier with the worst ranking. Among transformer models, the micro and macro F1 scores of TransformerXL and Longformer are better than the scores of the BERT-base and RoBERTa-base models. One of the main reasons for this is that BERT-base and RoBERTa-base models are restricted by the maximum sequence length of $512$ tokens, while TransformerXL and Longformer can handle longer sequences. Since MIMIC-III data are long with an average of $1,500$ tokens, the ability to encode more extended input data improves accuracy. However, the total time required for the experiments using TransformerXL and Longformer is exceptionally high. On the CD plot, a bold line connects PubMedBERT to the above two transformers indicating no statistically significant difference in average ranking. Compared to PubMedBERT, Longformer requires 5 times more processing time and TransformerXL almost 13 times more processing time. Among the embedding-based neural networks, CNNText and CAML are the best performing models, and they require less than an hour in total for an experiment. Micro-F1 scores of embeddings-based neural networks are better than those of most transformer models, except for TransformerXL.   

\begin{figure}[t!]

        \centering
        \includegraphics[width=0.95\textwidth]{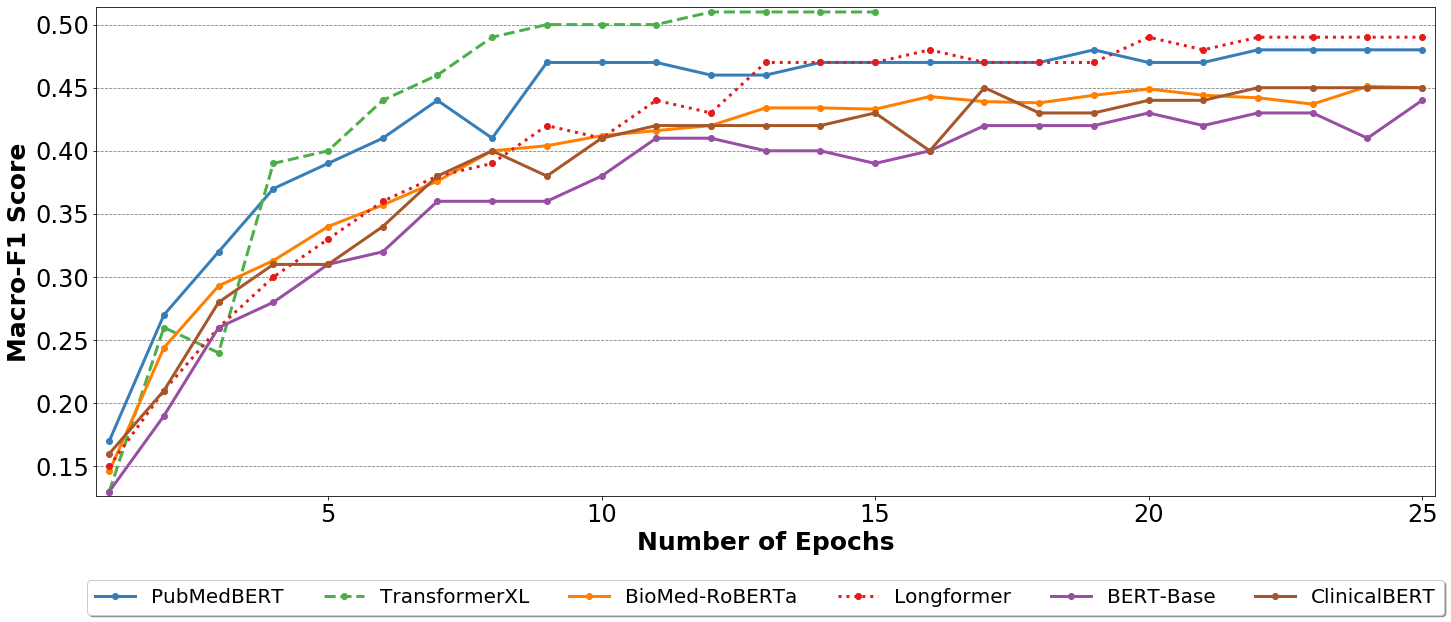}
     \caption{Macro F1 scores of transformer models for COVID-19 patient shielding over number of epochs for MIMIC-III data. Due to resource restrictions, TransformerXL was only run for 15 epochs. }
    \label{fig:covid_epoch}
    \vspace{-1em}
\end{figure}

\begin{figure}[ht!]
    \centering
     \includegraphics[width=\textwidth]{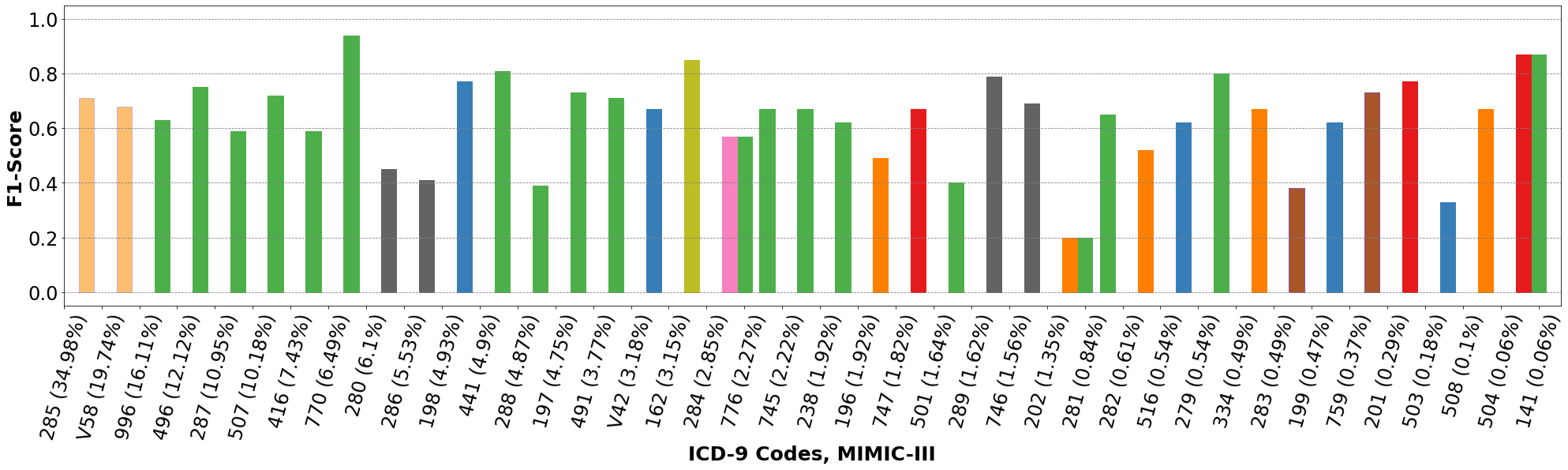}
      \includegraphics[width=\textwidth]{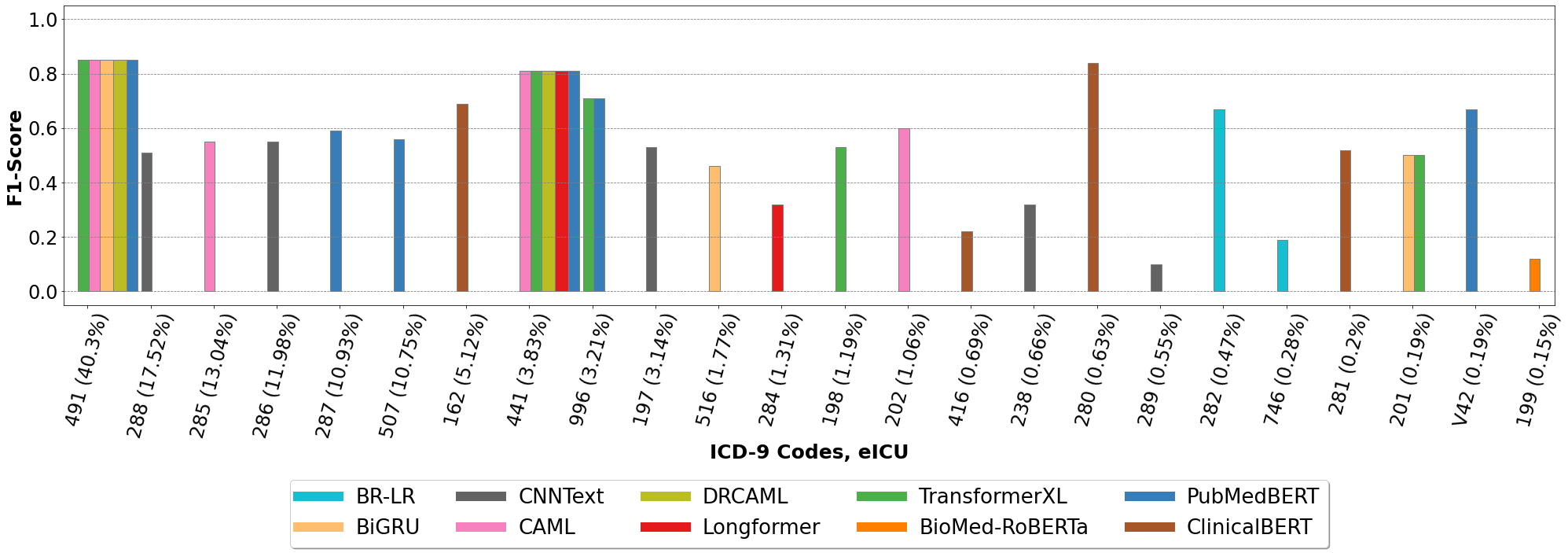}
    \caption{Best F1 scores and corresponding multi-label classifiers for individual labels with MIMIC-III (top) and eICU (bottom). Codes 235 and 142 for MIMIC-III data and  code 501 for eICU are omitted, as all models fail to predict them. }
    \label{fig:covid_f1_trans}
      \vspace{-1em}
\end{figure}

For eICU data, as observed in MIMIC-III, BR-LR and ECC-LR are the two worst-ranking classifiers in the CD-plot, with other classifiers performing better than ECC and BR overall. The micro-F1 score of PubMedBERT is the best among the multi-label classifiers, with a minimal difference in macro-F1 compared to the macro-F1 score of Longformer. TransformerXL requires 3.8 times more processing time than PubMedBERT and 71.6 times more processing time than CNNText. The Critical difference plot indicates no statistically significant difference between most classifiers, except BR-LR and ECC-LR (worst performers).

Figure~\ref{fig:covid_epoch} presents a comparison of macro F1 scores after each epoch for a range of 1 to 25 for transformer models presented in Table~\ref{tab:res_overall} for COVID-19 patient shielding using MIMIC-III data. For TransformerXL, the experiments were only performed for 15 epochs as the required computational resources for these experiments are very high.  
For macro-F1 scores, there is a steady increase in the number of epochs, where more are required before measures stabilise. For this research, macro-F1 scores were prioritised, as macro-F1 is an average over F1 scores of each label. In the case of predicting COVID-19 patient shielding, an increase in macro-F1 will indicate an increase in individual F1 score labels, where each label refers to a medical code associated with patient at high risk of COVID-19. The total training times presented in Table~\ref{tab:res_overall} are calculated for 20 epochs for all transformer models, except for TransformerXL. 

Figure~\ref{fig:covid_f1_trans} presents the best F1-scores for each label, and different colours are used to differentiate the corresponding classifiers. For MIMIC-III data, TransformerXL is the clear winner with the highest number of best F1 scores. For eICU data, PubMedBERT is marginally better than TransformerXL and CNNText. 


\section{Discussion}

There are many examples of the use of machine learning approaches in the fight against COVID-19. We propose a novel multi-label approach to predicting COVID-19 patient shielding from EHRs using publicly available information on COVID-19 and publicly available data. We present an extensive study on 12 different multi-label approaches to predicting medical codes where we include simple methods such as BR and ECC, examples of traditional word embeddings based neural networks, and the most recent transformer models. If overall predictive accuracy is the only deciding factor TransformerXL is the best option. However, given  pandemic situations where time is a significant factor, the predictive accuracy of models is not the only factor to consider in selecting an approach. 
To the best of our knowledge, this is the first study to consider such a range of multi-label text classification models to enable a better understanding of the potential to help fight against COVID-19.


\bibliographystyle{splncs04}
\bibliography{bib}

\end{document}